\documentclass[conference]{IEEEtran}
\IEEEoverridecommandlockouts
\usepackage{cite}
\usepackage{amsmath,amssymb,amsfonts}
\usepackage{algorithmic}
\usepackage{graphicx}
\usepackage{textcomp}
\usepackage{xcolor}
\usepackage{cuted}
\usepackage{float}
\usepackage{caption}
\usepackage{subcaption}
\usepackage{capt-of}
\usepackage[normalem]{ulem}
\usepackage{url}
\usepackage[T1]{fontenc}
\usepackage{flushend}
\usepackage[margin=0.75in]{geometry}
\def\BibTeX{{\rm B\kern-.05em{\sc i\kern-.025em b}\kern-.08em
    T\kern-.1667em\lower.7ex\hbox{E}\kern-.125emX}}
    
\usepackage{xcolor}
\newcommand{\cmt}[1]{}

\newcommand{\s}{\mathbf{s}}
\newcommand{\ac}{\mathbf{a}}

\long\def\ignorethis#1{}

\newcommand{\etal}{{\em{et~al.}\ }}

\newcommand{\vc}[1]{\ensuremath{\mathbf{#1}}}

\newcommand{\pctab}{\hspace{0.2in}}

\begin{document}
\newgeometry{left=0.75in, right=0.75in, top=1in, bottom=0.75in}
\title{\LARGE \bf Solving Challenging Control Problems via \\
Learning-based Motion Planning and Imitation
}

\author{
    Nitish Sontakke and Sehoon Ha
    \thanks{Georgia Institute of Technology, Atlanta, GA 30318 USA. e-mail: {\tt\scriptsize \{nitishsontakke,sehoonha\}@gatech.edu}}
}

\maketitle

\begin{abstract}
We present a deep reinforcement learning (deep RL) algorithm that consists of learning-based motion planning and imitation to tackle challenging control problems. Deep RL has been an effective tool for solving many high-dimensional continuous control problems, but it cannot effectively solve challenging problems with certain properties, such as sparse reward functions or sensitive dynamics. In this work, we propose an approach that decomposes the given problem into two deep RL stages: \emph{motion planning} and \emph{motion imitation}. The motion planning stage seeks to compute a feasible motion plan by leveraging the powerful planning capability of deep RL. Subsequently, the motion imitation stage learns a control policy that can imitate the given motion plan with realistic sensors and actuation models. This new formulation requires only a nominal added cost to the user because both stages require minimal changes to the original problem. We demonstrate that our approach can solve challenging control problems, \emph{rocket navigation}, and \emph{quadrupedal locomotion}, which cannot be solved by the monolithic deep RL formulation or the version with Probabilistic Roadmap. 
\end{abstract}


\section{Introduction}
Deep reinforcement learning (deep RL) has demonstrated impressive performance on a wide range of sequential decision problems ranging from the Go game \cite{silver2016mastering, silver2017mastering} to robot control~\cite{tan2018sim,haarnoja2018learning,hwangbo2019learning}. However, deep RL in practice still requires significant engineering efforts on tuning algorithms and reward functions to find an effective control policy. Conditions such as sparse reward functions and sensitive dynamics amplify the complexity of the problem. 
These properties degrade the accuracy of the policy gradient method by obscuring informative signals and causing inaccurate gradient estimation. Engineers often try to mitigate these issues by designing denser and smoother rewards or incorporating hand-designed controllers based on their prior knowledge.

Our approach to tackling challenging control problems is to divide them into easier sub-problems: \emph{motion planning} and \emph{motion imitation}. We solve planning with approximated dynamics inspired by other motion planning works~\cite{wermelinger2016navigation, belter2016adaptive, norby2020fast}. We then leverage the generated plan as a hint for solving the original problem. This decomposition gives substantial performance gains due to the following reasons. First, we can aggressively explore a wide range of states during the motion planning stage by leveraging direct sampling of states and cheap evaluation of transitions. We can also easily incorporate the desired prior knowledge in the planning, such as desired footfall patterns for locomotion tasks, which is not guaranteed in physics-based simulation. Although this planning might initially sacrifice minor details in physics, the imitation phase eventually recovers all the detailed dynamics and establishes robust control policies.

\begin{figure}
    \centering
    \includegraphics[width=0.81\linewidth]{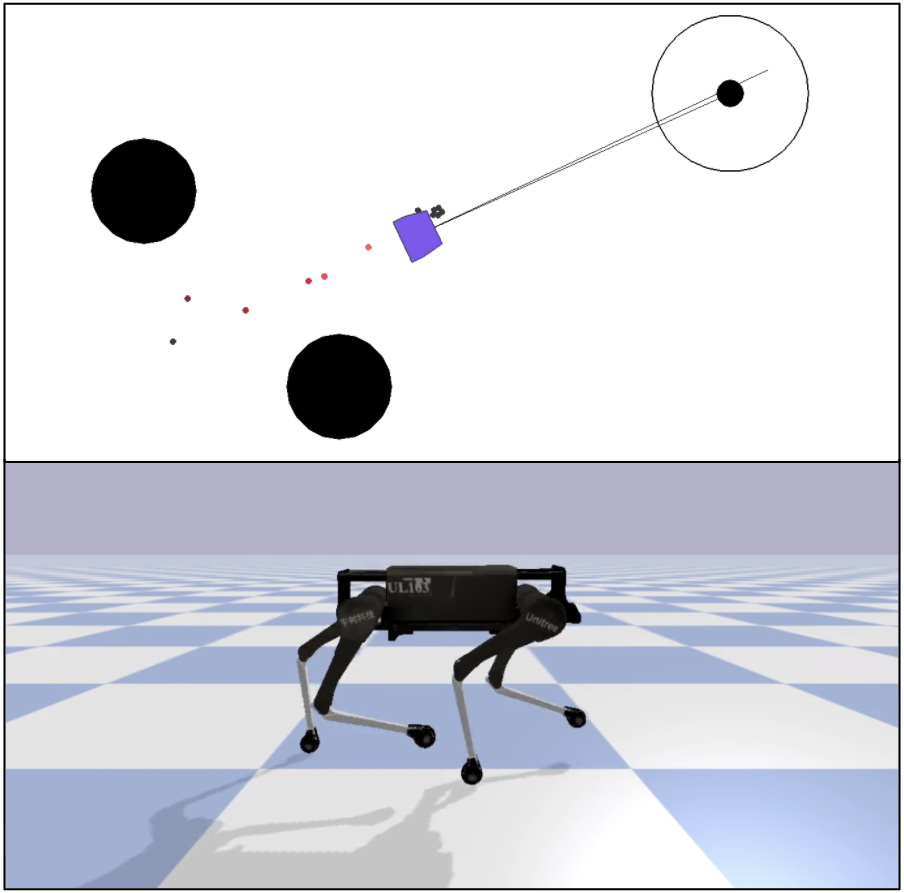}
    \caption{Two challenging environments used for evaluation: \emph{Rocket Navigation} and \emph{Quadrupedal Locomotion}.}
    \vspace{-0.5cm}
    \label{fig:teaser}
\end{figure}

Another crucial insight is to solve the motion planning via deep RL. Motion planning algorithms, such as Probabilistic Roadmap (PRM) \cite{kavraki1996probabilistic} or Rapidly exploring Random Tree (RRT) \cite{lavalle1998rapidly}, have been studied in various contexts including navigation and manipulation. However, these algorithms are not very effective in solving high-dimensional planning in our experience. On the other hand, we found that deep RL is an effective tool to solve complex planning problems by strategically exploring states via value function estimation, with minimal changes to the original control problem.

Our two-stage deep RL formulation works as follows. First, we solve the motion planning problem with the same reward function as the original problem but with a simplified transition function. We strategically explore various trajectories by sampling in the state space rather than sampling action signals. We suppress physics errors by introducing a new dynamics term, which is similar to the formulation in traditional motion planning algorithms. Once we generate the reference trajectory, we train a new motion imitation policy to track the given motion plan in a full-scale physics simulation. 

We evaluate the proposed algorithm in two environments: Rocket Navigation and Quadrupedal Locomotion, where both tasks are designed to have sparse reward functions and non-linear, sensitive dynamics. The experimental results indicate that our two-staged approach can achieve much higher rewards than the standard monolithic RL formulation or the version with PRM \cite{kavraki1996probabilistic}. We also demonstrate the robustness of the resulting policies by applying random perturbations at testing time.

\section{Related Work}

\noindent \textbf{Motion Planning Algorithms.}
Traditional motion planning encompasses various algorithms such as roadmaps, cell decomposition, potential field methods, and sampling-based methods. For domains such as quadrupedal locomotion, sampling tends to be the method of choice because of its efficacy in high-dimensional spaces. The Rapidly-Exploring Random Tree with its variants has gained significant traction as an effective tool for tackling this problem \cite{wermelinger2016navigation, belter2016adaptive, norby2020fast}. Ratliff \etal proposed CHOMP \cite{ratliff2009chomp}, which uses signed distance fields to model obstacles and Hamiltonian Monte Carlo Sampling instead of traditional sampling-based planning methods. There has also been a string of optimization-based methods~\cite{megaro2015interactive, farshidian2017efficient, fankhauser2018robust, carius2019trajectory, mastalli2020motion, ding2020kinodynamic}. Instead of using optimization, we solve motion planning problems using deep RL.

\noindent \textbf{Deep Reinforcement Learning.}
Learning-based approaches have proven to be effective for solving various control problems, including control of quadrupedal robots. Some of the earliest work to leverage reinforcement learning is by Kohl and Stone~\cite{kohl2004policy}, who proposed a policy gradient-based method to achieve a  faster gait on the  Sony Aibo robot. More recently, deep RL has been employed to train agents to learn feedback control~\cite{tan2018sim,pmtg,hwangbo2019learning}. The learned policy has been successfully deployed on real robots by using domain randomization (DR)~\cite{tan2018sim}, system identification~\cite{hwangbo2019learning}, or real-world adaptation~\cite{yang2020data}. Alternatively, researchers~\cite{haarnoja2018learning,ha2020learning} have investigated learning policies directly from real-world experience, which can intrinsically overcome sim-to-real gaps. Sample efficiency is a critical challenge for deep RL approaches, which can be improved by leveraging model-based control strategies~\cite{pmtg,tsounis2020deepgait,xie2021glide}. Recently, Lee \etal \cite{lee2020learning} employed a teacher-student framework for training their agent.
In addition, deep RL can solve a motion imitation problem~\cite{peng2018deepmimic,peng2020learning} to track the given reference motion, typically obtained by motion capture. Instead of using motion capture data, our algorithm aims to imitate the given motion trajectory obtained at the planning stage.

\noindent \textbf{Hierarchical Reinforcement Learning.} Hierarchical Reinforcement Learning (HRL)~\cite{jain2019hierarchical, jain2020pixels} has also proven to be an effective tool in tackling the locomotion problem.
Furthermore, researchers~\cite{peng2017deeploco} have developed HRL strategies for controlling legged characters in simulation, which have been further combined with adversarial learning~\cite{luo2020carl} to achieve high-level control.

\noindent \textbf{Reinforcement Learning with Motion Planning.} 
Traditional model-free RL algorithms are effective for many domains, but they often suffer from high sample complexity. They also have difficulty converging to optimal policies in the case of long-horizon tasks, and encounter safety concerns in environments cluttered with obstacles. On the other hand, motion planners need accurate models of the environment and are vulnerable to the curse of dimensionality. Their strength lies in their capability to compute optimal collision-free trajectories. 

Therefore, there exists a lot of prior work to combine the best of both domains. Jurgenson and Tamar~\cite{jurgenson2019harnessing} propose a modification to the popular DDPG algorithm that allows them to train neural motion planners more efficiently. Luo \etal~\cite{luo2021self} create demonstrations by planning to states already visited by the policy during the exploration phase. They train their policy by sampling experience tuples from this demonstration buffer and the original replay buffer. Xia \etal \cite{xia2020relmogen} propose a two-level hierarchical approach, where the higher-level module is a learned policy that outputs the sub-goal that the lower-level motion planner can then reach. Yamada \etal propose MoPA-RL \cite{yamada2020motion}, which switches between direct execution of policy actions and a path output by a motion planner, depending on the action magnitude. Faust \etal~\cite{faust2018prm} present PRM-RL, which utilizes probabilistic roadmaps with reinforcement learning agents as their local planners, allowing them to effectively use both techniques to address the shortcomings of each other.
Our approach draws inspiration from the given concept but instead uses deep RL for the motion planning component and combines it with the motion imitation framework.

\section{Two-staged Reinforcement Learning}
\begin{figure*}
    \centering
    \includegraphics[width=0.9\linewidth]{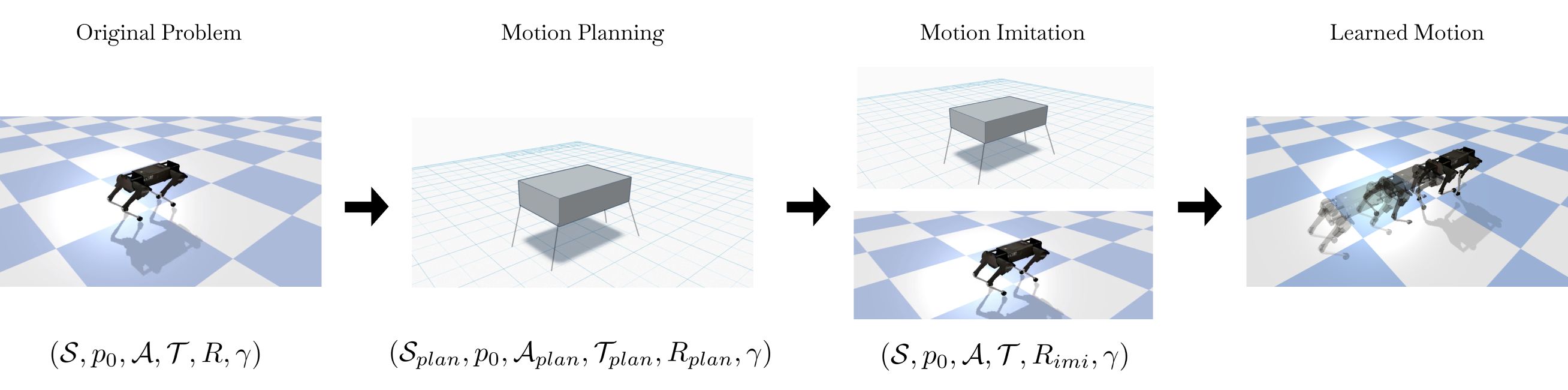}
    \caption{Overview of our two-staged Reinforcement Learning method. We decompose the problem into two deep RL sub-problems - motion planning and motion imitation. Motion planning learns an approximate motion plan using a simplified model. Motion imitation learns a control policy to track the generated reference plan in physics simulation.}
    \label{fig:overview}
\end{figure*}
Our algorithm solves challenging control problems by decomposing it into two deep RL stages: a \emph{motion planning} stage and a \emph{motion imitation} stage (Figure~\ref{fig:overview}).  During the \emph{motion planning} stage, we aim to find an approximate motion plan for the given problem using simplified dynamics. During the \emph{motion imitation} stage, we train a control policy with the original action space to imitate the given motion plan. 

\subsection{Background: MDP and Deep RL}
In reinforcement learning, we formulate the given problem as a Markov Decision Process (MDP) $(\mathcal{S}, p_0, \mathcal{A}, \mathcal{T}, R,  \gamma)$. We define an MDP with a set of states $\mathcal{S}$,  an initial state distribution  $p_0$, a set of actions $\mathcal{A}$,  a transition function $\mathcal{T} = p(\s_{t + 1}| \s_t, \ac_t)$, a reward function $R(\s_t, \ac_t)$, and a discount factor $\gamma$, where the subscript $t$ denotes the time step.

For each time step, an agent observes the current state $\s_t$ and samples an action $\ac_t$ from a policy $\pi_{\theta}(\ac_t | \s_t)$ that is parametrized by $\theta$. Our goal is to find the optimal policy parameter $\theta^{*}$ that maximizes the cumulative reward:
\begin{equation} \label{eq:1}
    J(\pi) = \mathbb{E}_{\tau \sim \rho_{\pi}(\tau)} \left[ \sum_{t=0}^{T} \gamma^t r_t \right],
\end{equation}
where $T$ is the planning horizon. Typically, deep RL utilizes a policy gradient that is estimated from a large number of trajectories to find the optimal parameters. This defines one of our baselines, monolithic RL, that solves the given problem  using a standard, unmodified deep RL algorithm such as Proximal Policy Optimization (PPO)~\cite{schulman2017proximal}.


However, deep RL cannot easily solve a challenging control problem when it is modeled using a standard MDP formulation. First, it often fails to solve a sparse reward function. Typically, it is easier to describe the desired behavior of the agent using sparse rewards by simply specifying the success condition or the final desired state. However, this easily confuses the deep RL algorithm because it does not receive informative signals during exploration. The second challenge is when we need to deal with high-dimensional, sensitive dynamics. For instance, in case of the bipedal locomotion problem, the majority of control signals will cause the robot to fall immediately. In this case, the control policy must be carefully tuned to generate meaningful experiences. To mitigate these challenges, we propose a two-stage approach that uses deep RL for the motion planning stage as well.

\subsection{Learning-based Motion Planning Stage} \label{sec:planning}
We re-utilize the deep RL approach, retaining the original problem definition as much as possible. In addition, using deep RL allows us to effectively solve the motion planning problem with an unknown time horizon, which is necessary for typical optimization-based motion planning algorithms.

We define a motion planning problem as a new MDP $(\mathcal{S}, p_0, \mathcal{A}_{plan}, \mathcal{T}_{plan}, R_{plan},  \gamma)$ with four new components: the state space $\mathcal{S}_{plan}$, the action space $\mathcal{A}_{plan}$, the transition function $\mathcal{T}_{plan}$, and the reward function $R_{plan}$. Note that $\mathcal{S}_{plan}$ and $R_{plan}$ require new formulations, while we can easily define $\mathcal{A}_{plan}$ and $\mathcal{T}_{plan}$ from $\mathcal{S}_{plan}$.

First, we define a compact state space $\mathcal{S}_{plan}$ that is typically a subset of $\mathcal{S}$.
Then we redefine the action space to directly sample the state space, which allows the learning algorithm to sample more meaningful states than random action signals. We define our action to be the difference to the state, $\ac \in \mathcal{A}_{plan} = \Delta \mathcal{S}_{plan}$, while heuristically limiting the delta between two states using box constraints. This gives rise to the new transition function $\s_{t+1} = \mathcal{T}_{plan}(\s_t,  \ac_t) = \s_t + \ac_t$. The simplicity of the transition function allows us to take a much larger step in the motion planning stage, such as $0.1$ second, compared to the typical time steps of $0.002$ seconds to $0.0005$ seconds in physics-based simulation. In our experience, this gives us a significant performance gain in terms of wall clock time.

However, we need an additional regulation because this new action space only considers the smoothness of the trajectory while ignoring physics. Therefore, we use an augmented reward for motion planning, $R_{plan} = R + R_{dyn}$, where $R$ is the original task reward and $R_{dyn}$ is the dynamics term that incentivizes the physical validity of the state transitions. We can implement $R_{dyn}$ in various ways: for instance, we can adopt an inverted pendulum for legged robots. 

By solving the given planning problem, the algorithm finds the trajectory of the states without control signals. This reference trajectory $\bar\tau = (\bar\s_0, \bar\s_1, \cdots, \bar\s_T)$ is used as an input to the next stage.

\subsection{Motion Imitation Learning Stage}
Once we obtain a reasonable trajectory,  our motion imitation stage is used to find a control policy that can generate feasible and effective control signals to follow the given trajectory. Once again, we define a motion imitation MDP $(\mathcal{S}, p_0, \mathcal{A}, \mathcal{T}, R_{imi},  \gamma)$ by only replacing the reward function $R$ of the original MDP with the motion imitation reward $R_{imi}$. The reward $R_{imi} (\s_t) $ is designed to measure the similarity between the current state and given reference state $\bar\s_t$. A typical choice would be to measure the difference between them, $R_{imi}(\s_t) = |\s_t - \bar\s_t|^2$. For more complex robots, such as quadruped robots, we can design a slightly more complicated reward function that measures the differences of various features, such as joint positions, joint velocities, and end-effector positions, inspired by the formulation in Peng et al.~\cite{peng2018deepmimic}. We will describe more details of our formulation in the experiments section.

Note that our two-staged approach allows us to convert sparse reward signals $R$ to dense and informative rewards, $R_{imi}$, with minimal prior knowledge. This invention allows us to find a control policy more efficiently. In addition, the motion imitation framework is known to find a robust control policy by repetitively collecting a large number of samples near the reference trajectory. We will further evaluate the benefit of our approach in the following section.

\section{Experiments}
We designed our experiments to evaluate the proposed two-staged learning framework on difficult control problems. Particularly, we selected two challenging problems, \emph{Rocket Navigation} and \emph{Quadrupedal Locomotion} (Figure~\ref{fig:teaser}), to answer the following research questions:
\begin{itemize}
\item Can the proposed method solve challenging control problems in a more sample efficient manner?
\item Can the proposed method allow us to incorporate prior knowledge into a control policy?
\item Is the resulting policy robust to external perturbations?
\end{itemize}
For all the learning experiments, we use the stable-baselines \cite{stable-baselines} implementation of PPO \cite{schulman2017proximal} as our learning algorithm. The hyperparameters for all the experiments, such as learning weights or network architectures, are tuned with a wide range using Bayesian hyperparameter search.

We compare our algorithms (\emph{Ours}) against two baselines. The first is a standard monolithic formulation (\emph{Monolithic}) that solves the given original MDP as-is. The second baseline is to solve the motion planning using the probabilistic roadmap (\emph{PRM}) algorithm and imitate the generated motion plan. Roughly, probabilistic roadmap randomly samples the state space and generates edges using a local planner. We construct the roadmap using Halton sampling to fill the state space with enough samples. We use a kd-tree for nearest neighbor computations, and employ Dijkstra's algorithm to obtain the reference trajectory. Once we obtain the trajectory, it is used as input to the subsequently defined motion imitation stage. Please refer to the original paper~\cite{kavraki1996probabilistic} for more details.



\subsection{Rocket Navigation}

Our first environment is \emph{Rocket Navigation}, where the task is to control the rocket so that it can reach the target location, the moon. This environment is inspired by the OpenAI gym~\cite{brockman2016openai} Lunar Lander environment and implemented using an open-source 2D physics simulator, PyBox2d \cite{catto2012pybox2d} with a frequency of $50$~Hz. The rocket is controlled by three engines, a single main engine responsible for thrust, and two side engines responsible for rotations. All the engines have highly non-linear relationships between control signals and actual exerted forces. We use a gravity value of $-2.5 m/s^2$, and model air resistance using the relationship $F_{ar} = -\lambda \Dot{\mathbf{x}}$, where $\Dot{\mathbf{x}}$ is the rocket velocity, and the coefficient $\lambda = 2.5$ in our experiments.

\noindent \textbf{Problem Formulation.} In the original problem, the observation is defined as a $10$ dimensional vector that includes the rocket's global position, orientation, velocity, angular velocity, relative location of the goal, current distance from the goal, and angle difference between the current rocket heading and goal direction. We use a two-dimensional continuous action space with values in the range $[-1, 1]$. The two dimensions correspond to the main and side engine thrusts respectively. The main engine only works when the action sampled is in $[0.5, 1.0]$. The side engines work when the action sampled is in the range $[-1.0, -0.5]\ \bigcup\ [0.5, 1.0]$: a negative value leads to clock-wise rotation whereas a positive value leads to counter-clockwise rotation. We scale our action values by factors $50$ and $10$ for the main and side engines respectively so that the rocket has enough thrust to complete the task successfully. 

\begin{figure}
\centering
\begin{subfigure}{.5\linewidth}
  \centering
  \includegraphics[width=\linewidth]{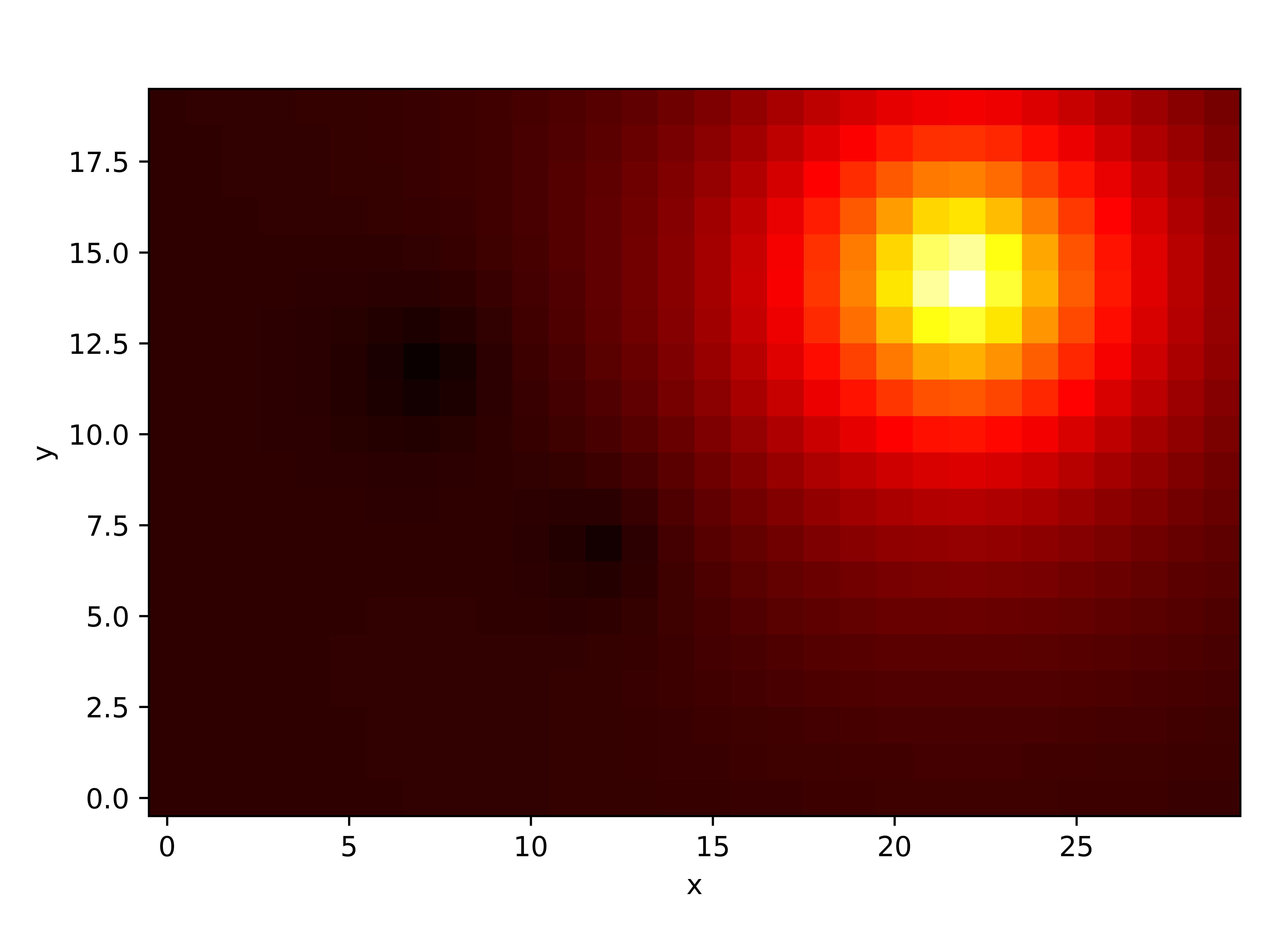}
\end{subfigure}%
\hfill
\begin{subfigure}{.5\linewidth}
  \centering
  \includegraphics[width=\linewidth]{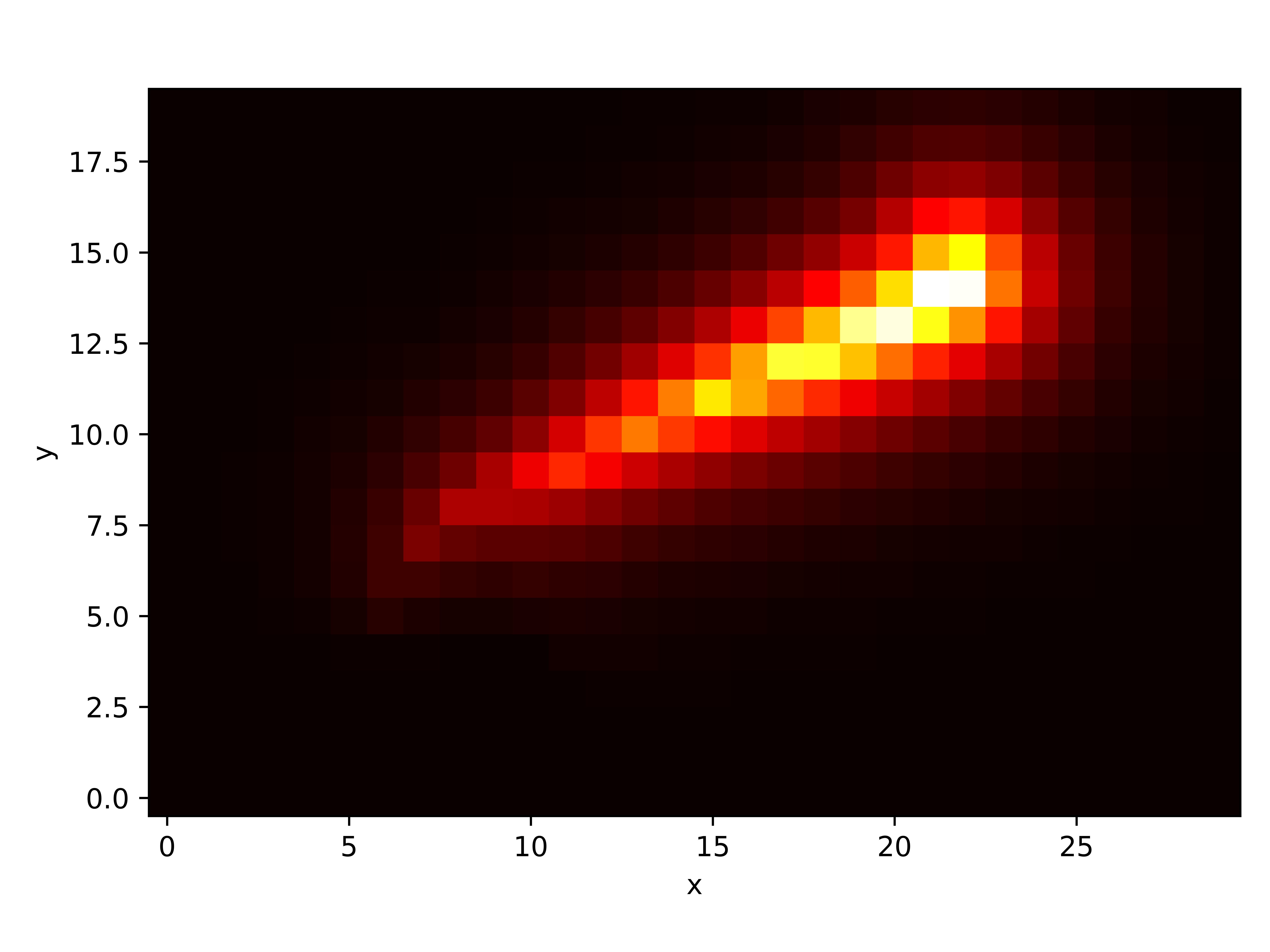}
\end{subfigure}
\caption{Heat maps of the reward functions in Rocket Navigation.
    \textbf{Left:} An original sparse reward $R$.
    \textbf{Right:} A dense imitation reward $R_{imi}$ to track the given reference motion.}
\label{fig:heatmap}
\end{figure}

For reward design, we use an exponential function that gives us explicit control over the smoothness of the reward:
\begin{multline*}
R(\s_t) = w_1\text{exp}(k_1||\vc{x}_t - \vc{x_g}||_2) \\ - w_2\text{exp}(k_2||\vc{x}_t - \vc{o_1}||_2) - w_3 \text{exp}(k_3||\vc{x}_t - \vc{o_2}||_2). 
\end{multline*}
Here, $\vc{x}_t$ is the current rocket position, $\vc{x_g}$ is the goal location, and $\vc{o_1}$ and $\vc{o_2}$ are the obstacle locations. The first term encourages the agent to get closer to the goal while the other two terms discourage collisions with the obstacles.  In our experiments, we set $w_1 = 0.9, w_2 = w_3 = 0.075, k_1 = -0.25, k_2 = k_3 = -0.5$.
Note that we intentionally design a sparse reward function by adjusting decaying factors: $k_1$, $k_2$, and $k_3$. The agent receives rewards and penalties only when the rocket gets very close to the goal or obstacles and does not receive informative signals if it is far from the goal.
We depict our sparse reward function using heat maps in Figure \ref{fig:heatmap} Left. This results in a challenging control problem that is difficult to solve using a naive RL algorithm.
We terminate the episode when (1) the rocket goes out of bounds, or (2) the episode reaches the maximum number of steps.


\noindent \textbf{Motion Planning MDP.} During the motion planning stage, we aim to find a sparse state trajectory that runs at $20$ Hz. We use a seven dimensional state space $\mathcal{S}_{plan}$ that is a subset of the original formulation without the velocities. The action $\ac \in \mathcal{A}_{plan}$ can directly change the position and orientation of the rocket by a maximum of $0.25$~m and $0.05^\circ$ at each step, which also naturally defines the transition function $\mathcal{T}_{plan}$ These values are chosen such that the resulting trajectory is realizable for the motion imitation policy. The reward function $R_{plan}$ is the same as the original reward $R$ without any additional $R_{dyn}$, which is sufficient to generate smooth and physically realizable trajectories. Note that we define the Motion Planning MDP by modifying the original MDP in a straightforward fashion.

\noindent \textbf{Motion Imitation MDP.} During the motion imitation stage, we find a control policy that imitates the reference trajectory generated by the motion planning stage. We only need to redefine the reward function to imitate the reference state:
$$R_{imi}(\s_t) = \text{exp}(k_{imi}||\vc{x}_t - \bar{\vc{x}}_t]||_2) R({\s}_t) - 0.25 I_c,$$
where $\vc{x}_t$ is the current rocket position and $\bar{\vc{x}}_t$ is the point on the reference trajectory closest to $\vc{x}_t$. $R$ is the original reward function.
The binary flag $I_c$ is set to $1$ if there is a collision, and $0$ otherwise. We set $k_{imi} = -0.625$. The heat map for this reward function is depicted on the right side in Figure \ref{fig:heatmap}. Here, the state space is the same as the baseline formulation.

\begin{figure}
    \centering
    \includegraphics[width=0.9\linewidth]{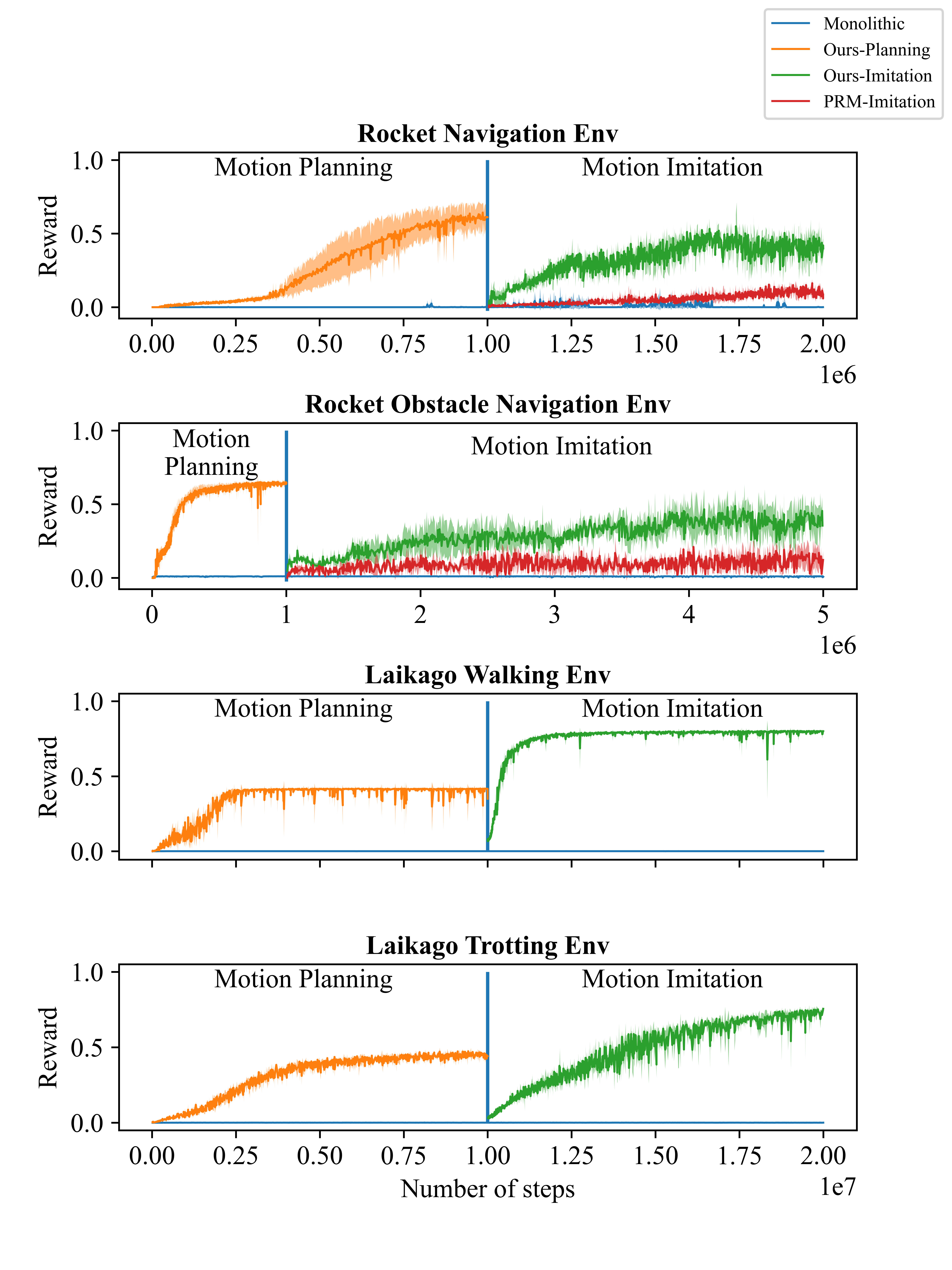}
    \caption{Learning curves for two Rocket Navigation and two Quadrupedal Locomotion environments. Our approach has two stages, motion planning (orange) and motion imitation (green), while the standard RL has one learning curve in blue. The rewards are normalized by their theoretical maximum value. We report our results as the average of three random seeds.}
    \label{fig:learning curves}
\end{figure}

\begin{figure}
\centering
\begin{subfigure}{.5\linewidth}
  \centering
  \includegraphics[width=\linewidth]{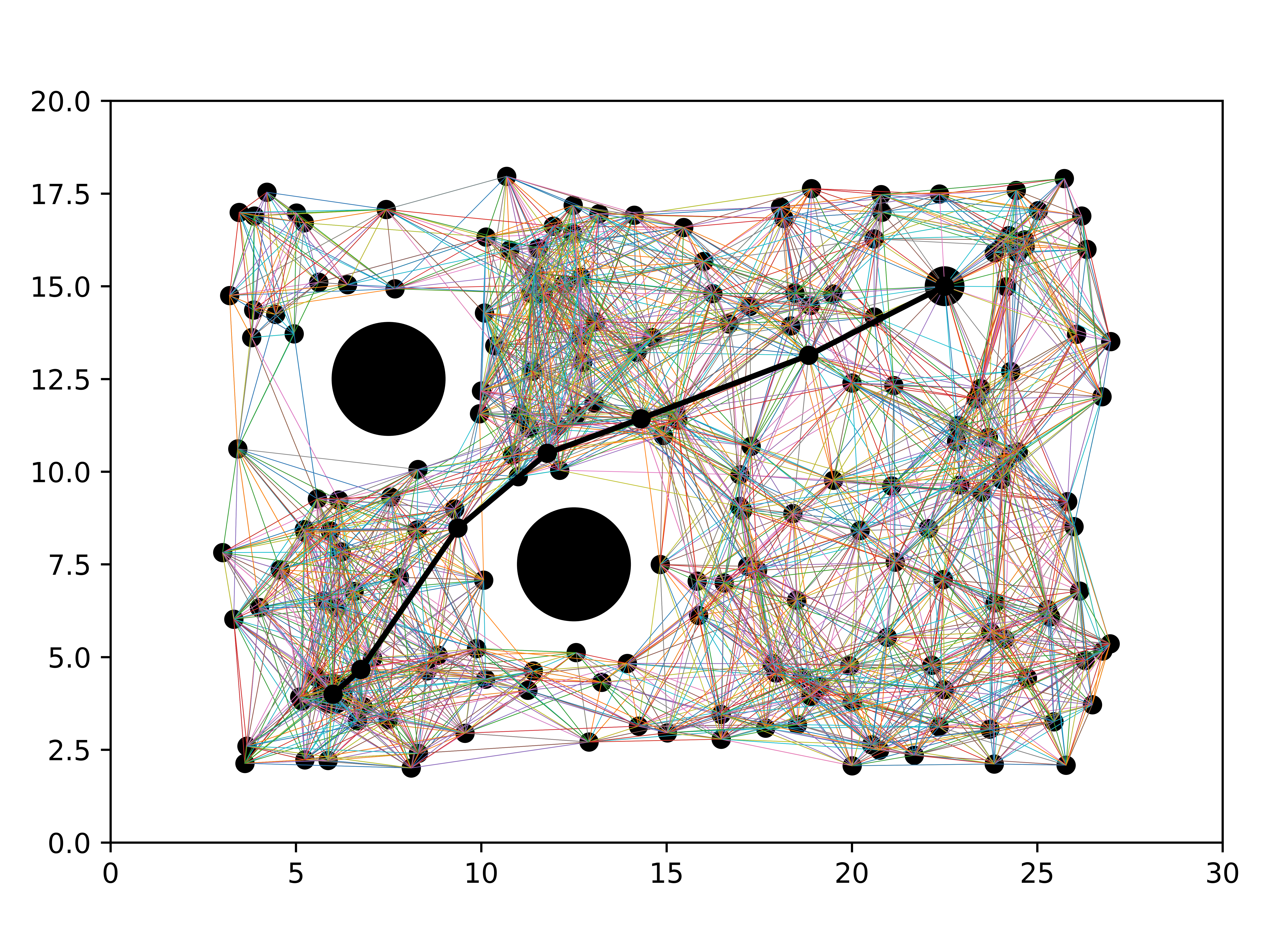}
  \caption{PRM}
  \label{fig:prm_ref_traj}
\end{subfigure}%
\hfill
\begin{subfigure}{.5\linewidth}
  \centering
  \includegraphics[width=\linewidth]{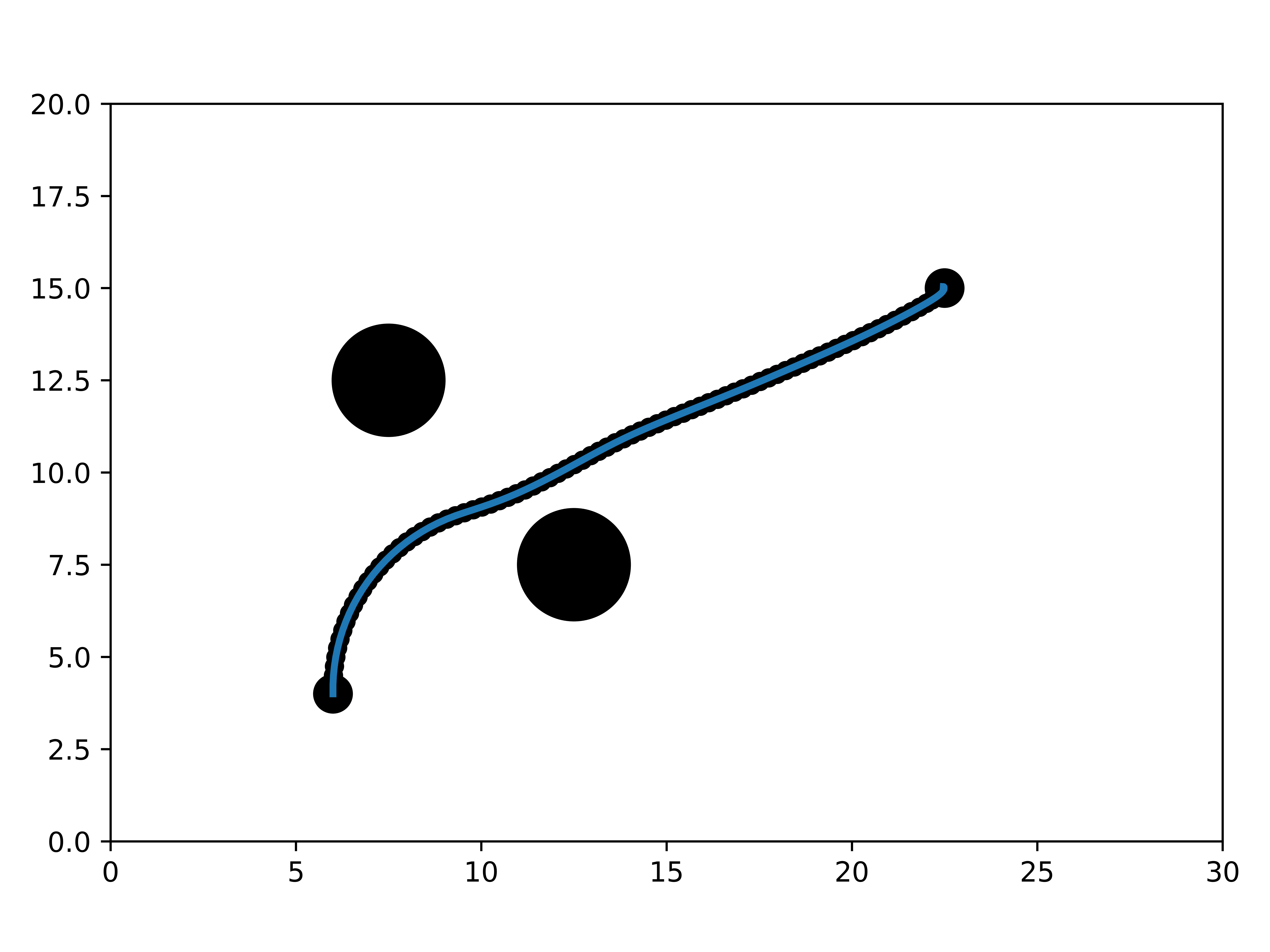}
  \caption{Ours-planning}
  \label{fig:ours_ref_traj}
\end{subfigure}
\caption{Comparison of the planned motion trajectories generated by PRM (Left) and our learning-based approach (Right).}
\label{fig:ref_traj}
\end{figure}


\noindent \textbf{Results.}
We investigated two scenarios, beginning with a simple navigation task, followed by the addition of two obstacles in the environment to make the task harder. For the simpler task, we omit the terms corresponding to the obstacles from the reward. In Figure~\ref{fig:learning curves}, we compared the learning curves of the proposed method (\emph{Ours}) against the baseline algorithms (\emph{Monolithic} and \emph{PRM}). We allowed one million time steps for our motion planning stage and one motion imitation stages for both \emph{Ours} and \emph{PRM}. The \emph{Monolithic} baseline is allowed two million time steps that is equal to the combined steps of \emph{Ours}. For harder environments, we allow five million steps for all the algorithms by allowing four million steps for the motion imitation stages.
Please note that one step in the motion planning stage is $5$ to $10$ times computationally cheaper for evaluation than the others. We compare normalized rewards for a fair comparison.

Both our method and \emph{PRM} were able to obtain effective solutions for both tasks, although the rewards of \emph{Ours} are slightly higher than those of \emph{PRM}. We believe that this difference arises from the difference in the smoothness of the reference trajectories, which can be observed in Figure~\ref{fig:ref_traj}. However, both \emph{Ours} and \emph{PRM} were able to obtain successful control policies that reach the goal without collisions. The \emph{Monolithic} baseline failed to solve either. We attribute this to the lack of meaningful reward signal received by the agent for moving toward the goal, owing to the sparsity of the reward. The \emph{Monolithic} agent only tried to explore nearby regions with random actuation signals and sinks to the bottom due to gravity. The simplified dynamics of the motion planning allowed the agent to explore a much larger region of the state space for both \emph{Ours} and \emph{PRM}.
We note that the reference motion produced by \emph{PRM} is similar to what \emph{Ours} generated. Although both learned paths are kinematic, we were able to learn a physics-based control policy by imitating the reference trajectory. 

The results support our key insight that decomposing the problem into the motion planning and imitation phases is effective: both \emph{Ours} and \emph{PRM} outperform \emph{Monolithic} by significant margins. Please note that we can obtain this performance gain by simply defining an intermediate motion planning problem that is straightforward to define.



\subsection{Quadrupedal Locomotion}
\begin{figure*}
    \centering
    \includegraphics[width=0.9\linewidth]{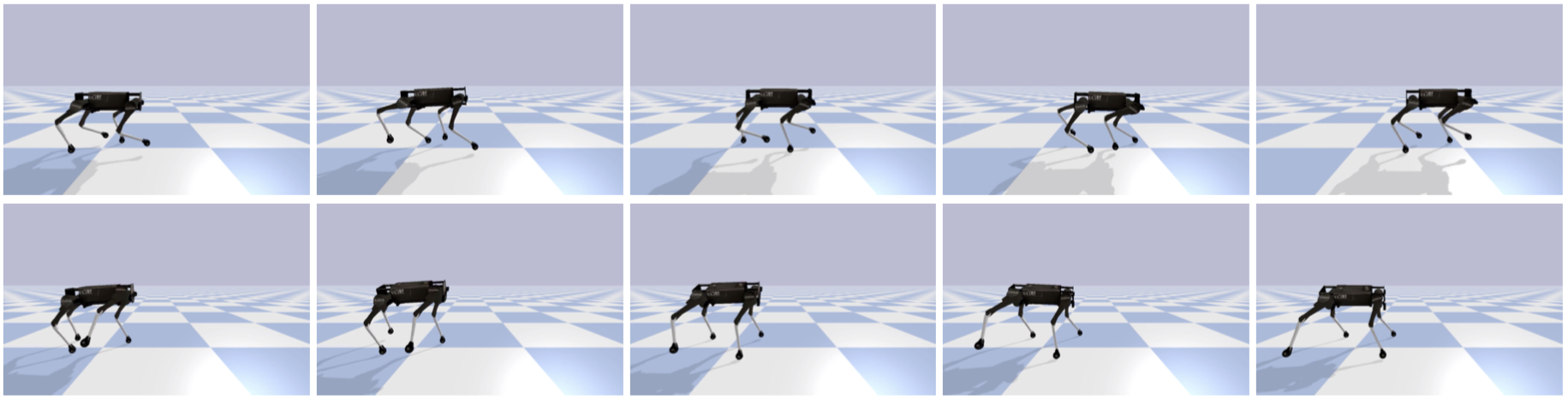}
    \caption{The learned Laikago trotting motions. Our two-staged approach finds a valid, periodic motion (\textbf{Top}) while the baseline barely balances without any forward movement (\textbf{Bottom}). The two motions are captured at the same time stamps.}
    \label{fig: laikago_motion_strip}
\end{figure*}
Next, we evaluate the method on a higher dimensional control problem, \emph{Quadrupedal Locomotion}. The task is to make the quadrupedal robot, Laikago from Unitree~\cite{laikago}, walk with a specified footfall pattern. We consider two gaits - walking and trotting. We utilize PyBullet~\cite{coumans2016pybullet} for simulation. We use a simulation frequency of $600$~Hz and a control frequency of $30$~Hz. Since the reward function does not include the detailed guidance for joint movements, it can still be considered sparse. In addition, the robot must maintain its balance during locomotion.
Therefore, this reactive control task is challenging to solve for legged robots with $12$ degrees of freedom without any additional prior knowledge, such as trajectory generators~\cite{pmtg}. 

\noindent \textbf{Problem Formulation.} In the original problem, the observation includes the robot's joint positions, joint velocities, bodies' linear and angular positions and velocities, a phase variable, and the distance to the goal. The 12-dimensional actions are the PD targets for all the joints. We define our multiplicative reward function as follows:
$$
R(\s_t, \ac_t) = r_{dist} \cdot r_{ns} \cdot r_{ee\_z}  \cdot r_{com}, 
$$
where the distance, no slip, end-effector ground z, foot clearance, and Center of Mass (CoM) terms respectively are 
\begin{align*}
    r_{dist} &= \text{exp} \left(k_{dist}||\vc{x}_t - \vc{g}||_2 \right) \\
    r_{ns} &= \text{exp} \left(k_{ns}\sum_{j=1}^{4} c(j, t) ||\vc{e(j)}_t - \vc{e(j)}_{t-1}||_2 \right) \\
    r_{ee\_z} &= \text{exp} \left(k_{ee\_z}\sum_{j=1}^{4} ||z(j)_t - \bar{z}(c(j, t))||_2 \right) \\
    r_{ee\_x} &= \text{exp} \left( k_{ee\_x} \sum_{j=1}^{4} ||x(j)_t - x(j)_{t-1}||_2 \right),    
\end{align*}
where the terms $r_{dist}$, $r_{ns}$, $r_{ee\_z}$, and $r_{ee\_x}$ are to minimize the distance to the goal pose, achieve the desired contact patterns, achieve the desired foot swing heights, and penalize the unnecessary foot movements. The term $\vc{x}_t$ is the current base position, $\vc{g}$ is the goal location, $c(j, t)$ is the desired contact flag for end-effector $j$ at time $t$, $\vc{e(j)}_t$ is its position, $z(j)_t$ its current z-coordinate and $\bar{z}(j)_t$ represents the desired z-coordinate which is $0$~cm when $c(j, t)$ is on and $10$~cm when off. 
We terminate the episode when the robot reaches the goal location, deviates from the goal location by more than the starting distance, its CoM drops below $0.32$m, or the episode reaches the maximum steps, which is set to $60$ in our experiments. We set $k_{dist} = -2.5, k_{ns} = -10, k_{ee\_z} = -20$ and $k_{ee\_x} = -4$ in our experiments.

\noindent \textbf{Motion Planning MDP.} We employ the abstract state space that consists of the body position and end-effector positions. Based on our definition in Section~\ref{sec:planning}, the action can directly change the kinematic positions of the abstract state. In addition, we directly control vertical positions of end-effectors based on the given footfall pattern: the foot height is set to $0$~cm if the foot should be in contact and $10$~cm if not. The base can move at a maximum velocity of $0.264$~m/s and $0.63$~m/s while the end-effectors can move at maximum angular speeds of $1.65$~rad/s and $1.26$~rad/s for walking and trotting respectively. 

We encourage the balance of the robot, which is omitted in the kinematic formulation, by introducing an additional pseudo-physics term, $R_{dyn}$:
$$R_{dyn} = \text{exp} \left(k_{com} || \vc{C} - \vc{P} ||_2 \right),$$
where it penalizes the deviation of the center of mass $\vc{C}$ from the center of the pressure $\vc{P}$, both projected onto the ground plane. With this term, we can define the motion planning reward $R_{plan} = R \cdot R_{dyn}$.


\noindent \textbf{Motion Imitation MDP.} During motion imitation, our goal is once again to find an imitation policy that can mimic the given reference motion from the previous stage. We use the same state and action spaces as the baseline and exactly the same reward function used in the original DeepMimic~\cite{peng2018deepmimic} formulation. Additionally, we use a Butterworth filter to smooth our actions. For more details, please refer to the original paper. 

\noindent \textbf{Results.}
As before, we compared the \emph{normalized} learning curves of our approach (\emph{Ours}) with the standard monolithic formulation (\emph{Monolithic}) in Figure~\ref{fig:learning curves}. We did not evaluate \emph{PRM} because it generates undesirable trajectories with irregular footsteps in the high-dimensional state space. We set the same budget of $10$M time steps for both stages of our method and $20$M time steps for the baseline. Similar to the previous rocket navigation environment, the monolithic baseline failed to learn meaningful locomotion policies. Our method could find an effective trajectory during the motion planning and converts it to a physically-valid control policy in simulation. Please refer to Figure~{\ref{fig: laikago_motion_strip}} and the supplemental video for the detailed motions.

Please note that our formulation has almost the same prior knowledge as the baseline, such as the footfall patterns or the desired speed. However, in our experience, the formulation makes a huge impact on learning results. In typical MDP formulations, the desired contact flags can only be encouraged by setting the reward term that has an indirect impact on learning. However, our framework allows us to directly couple them with the desired z positions of the feet.

\noindent \textbf{Robustness of Learned Policies.} In addition, the motion imitation framework is known to generate a robust control policy by repetitively sampling around the given target trajectory. We checked the robustness of the learned walking policy by applying external perturbations. The policy can endure up to the $100$~N force for a duration of $0.1$ second. We demonstrate the experiments in the supplemental video.

\subsection{Discussion about PRM and PRM-RL}
One of our key insights is to solve a motion planning problem using deep RL instead of traditional algorithms, such as PRM or RRT. We also tried these planning algorithms, but they did not perform well in challenging scenarios for the following reasons. First, many planning algorithms are vulnerable to high-dimensional state space. For instance, PRM begins by generating a lot of samples to fill the entire state space, which was not possible for our Laikago environments with six root joints and 12 rotational joints. We believe it is possible to search the state space more strategically by leveraging variants of those planning algorithms: for instance, RRT+~\cite{xanthidis2020motion}, RRT*-smart~\cite{islam2012rrt}, or A*-RRT~\cite{brunner2013hierarchical}. Hauser \etal\cite{hauser2008motion} developed an effective motion planning algorithm specialized for legged robots by combining it with a graph search. However, these methods typically require extra manual efforts in problem setup and hyper-parameter tuning. For instance, PRM can generate non-smooth trajectories if we populate too many samples. Additionally, determining edge connectivity in the graph while ensuring the existence of a feasible solution is also non-trivial. Owing to these challenges, we leave this further investigation as future work.

Instead, we simply generate a motion plan using standard deep RL algorithms, which also strategically search the space using learned value functions, with minimal modification to the original problem. In our experience, PRM works reasonably well in simple \emph{rocket navigation} environments while not being able to find good policies in \emph{quadrupedal locomotion} environments.

PRM-RL was also one of the baselines we considered. PRM-RL replaces a local planner with a deep RL learned policy to effectively define the connectivity between states. The authors demonstrated that PRM-RL effectively solves challenging real-world problems, such as long-range navigation in interactive environments. However, we found that it is not straightforward to apply PRM-RL to highly dynamic and sensitive control problems because we cannot decompose control into simple state-to-state transitions. In our experience, PRM-RL in \emph{rocket navigation} works poorly due to the dynamic nature of the environments.








\vspace{-5pt}

\section{Conclusion and Future Work}
We presented a deep reinforcement learning framework that takes a two-staged approach for solving challenging control problems. Our key idea is to split the given control problem into two sub-problems: \emph{motion planning} and \emph{motion imitation}. The motion planning phase is designed to quickly compute a feasible motion plan using approximated dynamics, while the motion imitation stage learns a control policy to imitate the generated motion plan. We evaluated the proposed algorithm on two challenging control problems, \emph{Rocket Navigation} and \emph{Quadruped Locomotion}, which have difficult reward functions and dynamics. The experimental results indicate that our two-staged approach can achieve much higher rewards than a standard RL formulation or a version with PRM.

Although the proposed framework is designed to solve general control problems, it still requires some domain-specific knowledge. For instance, the motion planning stage for the quadrupedal locomotion problem requires tuning the action representations and the additional approximated dynamics term, which can significantly affect the performance. In the future, we plan to investigate the proposed algorithm on more complex scenarios, such as locomotion with obstacles or manipulation. 

In this work, we studied the proposed algorithm on static environments with a fixed starting state. One interesting future research direction is to extend the motion planning phase to learn a resilient motion plan that can cope with dynamic perturbations by leveraging the full benefit of deep RL. For instance, we can learn a large number of motion plans for a quadrupedal robot that can deal with a wide range of external perturbations. This might require large-scale reinforcement learning to imitate many motor skills simultaneously.



\bibliographystyle{ieeetr}
\bibliography{refs}

\end{document}